# Interpreting Embedding Models of Knowledge Bases: A Pedagogical Approach

Arthur Colombini Gusmão [1]  Alvaro Henrique Chaim Correia [1]  Glauber De Bona [1]  Fabio Gagliardi Cozman [1]

## Abstract

Knowledge bases are employed in a variety of applications from natural language processing to semantic web search; alas, in practice their usefulness is hurt by their incompleteness. Embedding models attain state-of-the-art accuracy in knowledge base completion, but their predictions are notoriously hard to interpret. In this paper, we adapt "pedagogical approaches" (from the literature on neural networks) so as to interpret embedding models by extracting weighted Horn rules from them. We show how pedagogical approaches have to be adapted to take upon the large-scale relational aspects of knowledge bases and show experimentally their strengths and weaknesses.

## 1. Introduction

Large knowledge bases are now a reality (Bollacker et al., 2008; Miller, 1995), with applications from natural language processing to semantic web search (Schuhmacher & Ponzetto, 2014; Cucerzan, 2007). Even though a knowledge base may be quite sophisticated, in this paper we focus on knowledge bases that store triples of the form ⟨ head, relation, tail ⟩. Billions of such triples are now available in bases that range from specialized lexicons to broad repositories (Nickel et al., 2015). But even the largest knowledge bases are incomplete in the sense that they do not contain all triples that are true; several techniques aim at completing knowledge bases by automatically adding new triples.

The state-of-the-art in knowledge base completion typically relies on embedding models that map entities and relations into low-dimensional vector space. The existence of a triple is determined by some pre-defined function over these representations. More importantly, embedding models turn a complex space of semantic concepts into a smooth space where gradients can be calculated and followed.

One difficulty with embeddings is their poor interpretability. Of course, "interpretability" depends not only on a model, but also on its user. For instance, a seasoned statistician may find a very large logistic regressor to be easy to grasp. But embeddings seem particularly opaque as they turn a semantically rich input into numeric representations where each dimension bears little meaning. It is thus natural to think that we might interpret an embedding by translating it into a more interpretable model. However, this simple and promising idea requires caution when we deal with knowledge bases that are relational. First, we cannot rely on interpretable models that are solely propositional; rather, we must use relational constructs, such as Horn rules. Second, interpretability requires that we map back our decision making, even if they are built using values of the embedding, to the space of semantically-meaningful symbols.

The goal of this paper is to develop techniques to interpret embedding models associated with knowledge bases. We do so by adapting so-called "pedagogical approaches" that have been advanced in connection with shallow neural networks (Andrews et al., 1995). A pedagogical approach is one where, intuitively speaking, a non-interpretable but accurate model is run, and an interpretable model is learned from the output of the non-interpretable one.

Our idea is to extract weighted Horn rules from embedding models; these rules are both relational and interpretable, hence they satisfy our needs. We propose non-trivial changes that must be made to apply the pedagogical approach to our setting. In fact, there are several ways to apply pedagogical methods to interpret embedding models for knowledge bases; we discuss some of them, and show how to implement two of them. Finally, we present empirical results and discuss the properties of our methods.

The paper goes over needed background and related work in Section 2. In Section 3 we introduce our specific proposals, and in Section 4 we report experimental results. Section 5 closes the paper with a broad discussion.

[1] Escola Politécnica da Universidade de São Paulo, São Paulo, Brazil. Correspondence to: Arthur C. Gusmão ⟨arthurcgusmao@usp.br⟩, Fabio G. Cozman ⟨fgcozman@usp.br⟩.

## 2. Background and related work

In this section we offer a summary of important concepts in knowledge base (KB) completion (Section 2.1), and a brief review of relevant work on the interpretation of embedding models of knowledge bases (Section 2.2). In doing so we touch on issues that have been debated at length in previous work, such as the proper meaning and the quantification of interpretability (Doran et al., 2017; Lipton, 2016; Tintarev & Masthoff, 2007). In short, we assume that an interpretable system is one that allows the user to study and grasp its underlying mathematics (Doran et al., 2017).

### 2.1. Knowledge base completion

Let $\mathcal{E} = \{e_1, ..., e_{N_e}\}$ be the set of all entities and $\mathcal{R} = \{r_1, ..., r_{N_r}\}$ be the set of all relations in a KB $\mathcal{G}$, where $N_e$ and $N_r$ are the number of entities and relations, respectively. Each possible triple $x_{h,r,t} = \langle e_h, r_r, e_t \rangle$ can be modeled as a binary random variable $y_{h,r,t} \in \{0, 1\}$ that indicates its existence. Entity $e_h$ is the *head* and entity $e_t$ is the *tail* of $x_{h,r,t}$. Note that $y_{h,r,t} = 0$ does not imply that $x_{h,r,t}$ is false, but simply that it is not observed in $\mathcal{G}$.

The combination of all triples in a KB can be seen as a multigraph (called a *knowledge graph*) where nodes represent entities and directed edges represent relationships. An edge points from the head to the tail entity of a triple, while the type (or label) of an edge indicates its corresponding relation (Nickel et al., 2015).

There are two standard tasks in KB completion: *link (entity) prediction* and *triple classification*. The former consists of predicting a missing entity $e_t$ ($e_h$), given the relation $r_r$ and another entity $e_h$ ($e_t$) (Bordes et al., 2013). The latter consists of predicting the existence of a triple $\langle e_h, r_r, e_t \rangle$ (Socher et al., 2013). In this paper, we focus on triple classification and leave link prediction for future work.

**Embedding models** An *embedding model* represents entities and relations in a continuous vector space, and defines a score function $f(x_{h,r,t}; \Theta)$ that represents the plausibility that $x_{h,r,t}$ exists given the set of all parameters $\Theta$; usually one learns the representation of entities and relations by solving an optimization problem that maximizes the total plausibility of observed facts (Wang et al., 2017).

To perform triple classification, usually a threshold $\theta_r$ is found for each relation $r_r$ using a validation set. The triple $x_{h,r,t}$ is classified as present if $f(x_{h,r,t}) < \theta_r$, and classified as absent otherwise (Nguyen, 2017).

**Subgraph Feature Extraction** *Graph feature models* predict the existence of a triple by extracting features from edges in the knowledge graph (Nickel et al., 2015). Here we focus on a specific method of this class, namely, *subgraph feature extraction* (SFE) (Gardner & Mitchell, 2015), a variant of the *Path Ranking Algorithm* (PRA) (Lao & Cohen, 2010; Lao et al., 2011) shown to be faster and to achieve better performance. SFE extracts binary features from a graph that indicate the existence of a path (a set of relations) between two entities. For each relation, extracted features are saved into a *feature matrix*, later used with any desired classification model (typically logistic regression).

More formally, let $\pi$ denote a path type defined by some sequence of edges (relations) -$r_1$-$r_2$-...-$r_l$- in a knowledge graph $\mathcal{G}$. SFE constructs a subgraph $\mathcal{G}_n$ centered around each entity $e_n \in \mathcal{G}$ using $k$ random walks. Each random walk that leaves $e_n$ follows some path type $\pi_{n,i}$ and ends at an intermediade node $e_i$. To construct a PRA-like feature vector for a source-target pair $(e_n, e_m)$, SFE merges the subgraphs $\mathcal{G}_n$ and $\mathcal{G}_m$ on the intermediate nodes $e_i$, taking the combinations of all path types $\pi_{n,i}$ and $\pi_{m,i}$ for all $e_i$ as binary features. Each feature vector is saved as a row in a feature matrix, that can be used as an input to a classifier.

Because features extracted by PRA can be understood as bodies of weighted rules, the model is usually regarded as "easily interpretable" (Nickel et al., 2015). As features are restricted Horn clauses extracted from the graph, the method is closely connected to logical inference (Gardner et al., 2015). SFE also allows for extending its feature space and using more expressive features, but in this work we restrict ourselves to the PRA-like features mentioned above.

### 2.2. Interpreting embedding models: related work

The score functions of embedding models used in knowledge base completion can generally be viewed as neural networks (Nickel et al., 2015; Nguyen, 2017; Wang et al., 2017). Thus we start by reviewing concepts from the literature of non-relational rule extraction from neural networks. A considerable part of the literature on this topic comes from the 1990s. Andrews et al. (1995) consider *decompositional* techniques that focus on extracting rules from the level of hidden layers in the neural network (the internal structure of the network is seen as transparent), and *pedagogical* techniques that treat the network as a black box. Methods that combine elements of both approaches are called *eclectic*. Recent work can be found for both pedagogical (Augasta & Kathirvalavakumar, 2012) and decompositional (Zilke et al., 2016) approaches, with this last one extending a decompositional approach to deep networks.

In the relational setting, we find methods that are able to extract knowledge from neural networks in the form of rules. Techniques mentioned here are also reffered to as *neural-symbolic integration* (d'Avila Garcez et al., 2015). More recently, França et al. (2015) adapted TREPAN (Craven & Shavlik, 1996) to extract rules from a network trained in data propositionalized from a first-order example set. Srinivasan

Interpreting Embedding Models of Knowledge Bases& Vig (2017) construct explanations from a logical model that uses the structure and prediction of a network whose inputs are patterns in first-order logic.

There are also a variety of techniques that aim at obtaining interpretability in the context of KBs: Murphy et al. (2012) take the interpretability of each dimension of a word embedding as the capacity one has of distinguishing an intrusive word, i.e., one which has a low value in that dimension in comparison to other words in a group. From that perspective, they proposed a variant of matrix factorization that is highly interpretable. Chandrahas et al. (2017) proposed a technique for inducing interpretability in KB embeddings by incorporating additional entity co-occurence statistics from text, while still maintaining comparable performance in predictive tasks. Barbieri et al. (2014) proposed a stochastic topic model for link prediction that produces explanations based on the type of each predicted link (links can be "topical" or "social"). KSR (Xiao, 2016) learns semantic features for knowledge graphs. Xie et al. (2017) proposed an embedding model and designed a learning algorithm to induce interpretable sparse representations in it. Engelen et al. (2016) developed an efficient and explainable technique for link prediction using topological features.

Perhaps the previous effort that is closest to ours has been reported by Carmona & Riedel (2015). They have employed a pedagogical technique for extracting interpretable models from matrix factorization models fed with relational datasets. Their application is relational but avoids difficulties with feature building and instance generation that appear with large knowledge bases, issues we have tackled and whose solution alternatives are detailed in the next section.

## 3. Interpreting embedding models of knowledge bases

It is not trivial to apply either decompositional or pedagogical techniques for neural networks to relational embeddings. The main reason is that one cannot operate in the vector space in which the classification takes place; one must return to the space of entities and relations where interpretations can make sense to the user. How sensible is an explanation generated by a logistic regression operating on $\Re^{40}$, if it states that dimension 28 is the one determining a triple to exist? It is desirable for the interpretation to resort to entities and relations of the knowledge base, not to mapped values. We must find a way to generate instances in the semantic space of triples, in such a way that generated instances adequately capture the behavior of the embedding. From those generated instances we must learn relational rules.

The method proposed here does not restrict the model's structure or its input feature space. Thus, it differs from previous work by being the first that, to the best of our knowledge, interprets embedding models for knowledge base completion via weighted Horn rules in a model-agnostic fashion. We note that pedagogical approaches allow a model to have its relative trust assessed and compared to other methods, a property that is specially important in the scenario where the machine learning practitioner has to select a model among a number of alternatives (Ribeiro et al., 2016).

### 3.1. Explaining knowledge embedding models with predicted features (XKE-PRED)

The first method that we propose for explaining embedding models of KBs is the most direct application of the pedagogical approach we can think of. We treat the embedding model as a black box and assume no other source of information for building the interpretable model. By changing the original classifier's inputs and observing its outputs, the pedagogical approach constructs a training set for an interpretable classifier from which explanations are extracted. Each input to the original classifier consists of a triple that has no inherent interpretable features; to overcome this difficulty, we resort to applying SFE to the data generated by the black box. This way we can apply the default pedagogical framework: by using features extracted from SFE and labels predicted by the original classifier, we train a logistic regression model, from which we draw explanations in the form of weighted Horn clauses.

More formally, let $\mathcal{T} = \mathcal{E} \times \mathcal{R} \times \mathcal{E}$ denote the set of all possible triples in a relational setting; let $\Pi_\mathcal{G}$ represent the set of all possible path types between two entities in a knowledge graph $\mathcal{G}$ and $\mathcal{P}(\Pi_\mathcal{G})$ its power set; let $g \colon \mathcal{T} \mapsto \{0, 1\}$ denote the original, black box classifier; and let $F \colon \mathcal{T} \to \mathcal{P}(\Pi_\mathcal{G})$ denote the feature extraction function performed by SFE for a triple $x_{h,r,t} \in \mathcal{T}$, given $\mathcal{G}$. We call *explaining knowledge embedding models with predicted features* (XKE-PRED) the following scheme. First, construct a set of examples $\mathbb{D}' = \{(F(x_{h,r,t} \mid \hat{\mathcal{G}}), g(x_{h,r,t}))\}^n$ of arbitrary size $n$ for training an interpretable classifier $g' \colon \mathcal{P}(\Pi_{\hat{\mathcal{G}}}) \mapsto \{0, 1\}$, where $\hat{\mathcal{G}} = \{g(x_{h,r,t}) \mid x_{h,r,t} \in \mathcal{T}\}$ represents the graph composed of all triples predicted as correct by $g(\cdot)$. Then draw explanations from $g'(\cdot)$ as a function of $\mathcal{P}(\Pi_{\hat{\mathcal{G}}})$.

Generating $\hat{\mathcal{G}}$ by trying all possible triples is not feasible in practice, as $\|\mathcal{T}\| = \|\mathcal{E}\|^2 \cdot \|\mathcal{R}\|$ grows quadratically with the number of entities. Even for relatively small graphs such as the ones in Table 2 we already have billions of possible triples. Moreover, only a small subset of $\mathcal{T}$ is likely to be true (Nickel et al., 2015).

To arrive at a tractable number, a naïve approach would be to sample a subset of triples $\mathcal{T}_0$ from $\mathcal{T}$ and build a subset $\hat{\mathcal{G}}_0$ of $\hat{\mathcal{G}}$. We propose a more sophisticated method: after randomly sampling $\mathcal{T}_0$ of arbitrary size, we generate an extension $\mathcal{T}_0^k$ of it by corrupting both entities (head and

81

tail, simultaneously) for each triple predicted as correct in $\mathcal{T}_0$ with its $k$ nearest neighbors (including the entity itself). That is, given an arbitrarily sufficient number of triples $\mathcal{T}_0^+$ classified as correct, $\hat{\mathcal{G}}_0^k = \{g(x_{h',r,t'}) \mid x_{h,r,t} \in \mathcal{T}_0^+ \wedge e_{h'} \in N_k(e_h \mid \Theta_g) \wedge e_{t'} \in N_k(e_t \mid \Theta_g)\}$ represents the graph of correct triples inferred by the classifier $g(\cdot)$ with parameters $\Theta_g$, where $N_k(e \mid \Theta_g)$ is the set that contains $e$ and its $k-1$ nearest neighbors in the entities' vector representation given by $\Theta_g$.

The motivation for the method presented above is to generate the predicted graph as complete as possible (with as many positive examples as possible). Corrupting triples classified as correct by replacing their entities with their nearest neighbors improves the likelihood of finding novel positive instances. In practice, it is often the case that the machine learning practitioner has access to the embedding model's training data, which can be used as $\mathcal{T}_0$, since positive training triples tend to be classified as correct by the model. We note that the idea of using nearest neighbors to navigate the training data is not new in the pedagogical setting (Etchells & Lisboa, 2006).

An interesting characteristic of XKE-PRED is that explanations depend on other predictions (hence, its name). By looking at an explanation, one is analyzing not how features from the real world influence the model's predictions, but how the model organizes its internal representation. In other words, the method captures correlations between the model's predictions and analyzes their internal coherence. One must question whether this is the best way to interpret a prediction. One may be more interested in understanding how patterns from the real world influence the embedding's decisions, and not solely if the embedding's internal representations make sense. To handle that, we propose another method in the next section, that allows for the usage of an external source of interpretable features in explanations.

### 3.2. Explaining knowledge embedding models with observed features (XKE-TRUE)

In this section we propose a variation of XKE-PRED that assumes an *external source* of knowledge besides the embedding model, regarded as *ground truth* of our relational domain, from which we extract interpretable features. We call this approach *explaining knowledge embedding models with observed features* (XKE-TRUE). The motivation behind it is that we want to explain the black box's predictions based on *real* features, instead of predicted ones, therefore we use "TRUE" in its acronym.

Following the notation presented in Section 3.1, let $\mathcal{G}$ represent the ground truth for a relational setting, acquired from an external source. XKE-TRUE constructs a set of examples $\mathbb{D} = \{(F(x_{h,r,t} \mid \mathcal{G}), g(x_{h,r,t}))\}^n$ of arbitrary size $n$ for training an interpretable classifier $g'' : \mathcal{P}(\Pi_\mathcal{G}) \mapsto \{0, 1\}$, from which we draw explanations that are a function of $\mathcal{P}(\Pi_\mathcal{G})$. The main difference of this method is that $\mathcal{G}$ contains information about a set of instances, and, therefore, we are explaining each prediction from the embedding model with a set of observations from the real world.

In practice, it may be interesting to use all available data for both training the embedding model and extracting features for XKE-TRUE to maximize the amount of information at each stage (the method does not require such action).

## 4. Experiments

Here we present and compare experimental results for both XKE variants (code available[1]). The datasets used in our experiments are described in Table 2. For each dataset we trained TransE (Bordes et al., 2013), a simple yet efficient embedding model that is generally used as baseline for knowledge base completion (Guo et al., 2015; Trouillon et al., 2016; Liu et al., 2017). The TransE model is inspired by methods such as Word2Vec (Mikolov et al., 2013) as it represents entities and relations as points in the same vector space. Relationships are represented as translations in the embedding space: if a triple $x_{h,r,t}$ holds, the vector that represents the tail entity $\vec{e_t}$ should be close to the head entity vector $\vec{e_h}$ plus the relation vector $\vec{r_r}$.

### 4.1. Evaluation criteria

To measure the quality of the weighted rules, we use metrics that reveal performance and interpretability. The performance metrics are *fidelity* and *accuracy*. Fidelity (ratio of prediction matches) defines the ability of the pedagogical method to mimic the behavior of the embedding model; accuracy defines the weighted rules' ability of correctly predicting real data.

For interpretability, we report the mean number of rules (features with weight greater than zero) and the mean rule length (number of relations in each path). These are objective measures that ignore the subjective nature of interpretability (Freitas, 2013); we also offer a qualitative analysis that is of a more subjective character.

### 4.2. Model training

The embedding models were trained via grid search following Nguyen et al. (2016). Negative examples were generated using the Bernoulli distribution procedure introduced by Wang et al. (2014), and training was limited to 1000 epochs.

XKE-PRED and XKE-TRUE were applied following the procedure described in Section 3. The same data used for training TransE was also used as the external source of information for XKE-TRUE and for initializing $\mathcal{T}_0$ in XKE-PRED. We extracted features using SFE, with the maximum rule length limited to four due to performance reasons, and trained a logistic regression model for each relation. Negative examples were generated via the Bernoulli procedure, with two negative examples for each positive one.

---
[1] https://github.com/arthurcgusmao/XKE

Table 1. Results (micro-average) for both XKE variants. XKE-PRED is indexed by the number of nearest neighbors used for generating $\hat{\mathcal{G}}$.

| Dataset | FB13 | | | | NELL186 | | | |
|---|---|---|---|---|---|---|---|---|
| XKE variant | TRUE | PRED$_3$ | PRED$_5$ | PRED$_7$ | TRUE | PRED$_3$ | PRED$_5$ | PRED$_7$ |
| Embedding Accuracy | 82.55 | | | | 86.40 | | | |
| # Positive triples in $\mathcal{G}$ (XKE-TRUE) or $\hat{\mathcal{G}}$ (XKE-PRED) | 322k | 830k | 1,668k | 2,658k | 36k | 196k | 524k | 987k |
| $\hat{\mathcal{G}}$ positive over predicted ratio | - | 0.286 | 0.207 | 0.168 | - | 0.604 | 0.581 | 0.558 |
| # Features per example | 2.91 | 0.91 | 1.34 | 1.79 | 70.66 | 159.54 | 249.86 | 337.41 |
| % Examples with # features $> 0$ | 54.73 | 33.83 | 37.88 | 41.81 | 50.01 | 39.39 | 45.57 | 51.87 |
| Explanation Mean # Rules (for explanations with size $> 0$) | 2.29 | 2.19 | 2.70 | 2.57 | 105.30 | 51.33 | 159.02 | 158.87 |
| Explanation Mean Rule Length | 3.09 | 3.00 | 2.87 | 2.82 | 3.86 | 3.78 | 3.89 | 3.89 |
| Fidelity | 73.26 | 66.65 | 74.36 | 69.99 | 86.55 | 77.00 | 74.94 | 75.64 |
| Fidelity (filtered for examples with # features $> 0$) | 80.52 | 84.30 | 85.74 | 83.28 | 87.02 | 85.00 | 83.07 | 84.47 |
| Fidelity (weighted by the # features) | 75.21 | 82.67 | 84.58 | 84.80 | 85.66 | 88.09 | 86.24 | 88.22 |
| Accuracy | 73.43 | 64.58 | 71.78 | 68.11 | 89.10 | 75.79 | 76.18 | 76.44 |
| Accuracy (filtered for examples with # features $> 0$) | 80.78 | 81.00 | 82.02 | 80.34 | 91.19 | 84.08 | 84.30 | 85.11 |
| Accuracy (weighted by the # features) | 71.68 | 78.42 | 81.28 | 82.19 | 82.12 | 86.56 | 89.11 | 89.41 |
| F1 (Fidelity) | 76.66 | 50.11 | 71.14 | 61.13 | 83.19 | 61.41 | 68.07 | 68.03 |
| F1 (Accuracy) | 77.35 | 49.07 | 69.16 | 59.69 | 86.89 | 62.66 | 71.14 | 70.68 |

Table 2. Datasets used in our experiments: FB13 is a subset of Freebase (Bollacker et al., 2008) introduced by Bordes et al. (2013). NELL186 a subset of NELL (Mitchell et al., 2015) introduced by Guo et al. (2015). Both datasets contain negative examples (not included in the count below) for validation and test.

| Dataset | $\|\mathcal{E}\|$ | $\|\mathcal{R}\|$ | $\|\mathcal{T}\|$ | Train | Valid | Test |
|---|---|---|---|---|---|---|
| FB13 | 75,043 | 13 | $73 \cdot 10^9$ | 316,232 | 5,908 | 23,733 |
| NELL186 | 14,463 | 186 | $39 \cdot 10^9$ | 31,134 | 5,000 | 5,000 |

### 4.3. Results

We present the results of both XKE-PRED and XKE-TRUE in Table 1. Regarding feature extraction, we see that the entity similarity method worked well to increase the size of $\hat{\mathcal{G}}$ and the number of extracted features per example. It did not, however, increase the percentage of examples with at least one extracted feature, which was relatively low in all cases. Without features, the explanation for a given example is defined only by the bias term of the logistic regression.

Considering interpretability, the mean rule length is less than four for all cases (by constraint), indicating that the Horn clauses whose weight is greater than zero are short, which we consider easily interpretable. The mean number of rules in FB13 is small, favoring interpretability, but it increases in NELL186, going up to almost 160.

Regarding fidelity, both methods achieve over 80% for examples with at least one feature (filtered results), but this number drops when we consider all examples. Also, there is little or no improvement comparing the metrics weighted by the number of features with the filtered ones. These results indicate that the probability of correctly classifying an example increases considerably from zero to one feature extracted, and, after that, the impact gets less relevant as we add more features; thus, we can understand why XKE-PRED had, on average, lower fidelity than XKE-TRUE. Initially, we would expect the former to achieve superior fidelity because it predicts the embedding model decisions using a graph defined by the embedding itself. The fact that we were not able to increase the percentage of examples with at least one feature, even when extracting them from graphs with 8 times the number of positive triples than the original graph, possibly indicates an internal inconsistency in the embedding model with regard to relational modeling.

For accuracy, the values achieved are very close to the corresponding fidelity value. When the embedding model's accuracy is higher than the interpretable model's accuracy, one may choose to use the embedding for prediction and the logit for explanation, depending on the accuracy-interpretability trade-off position one is willing to take. However, when the interpretable classifier's accuracy is higher than that of the black box, which happened for XKE-TRUE in NELL186, the best option is to choose the interpretable model for both predicting and explaining. In this case, the optimal procedure is to use labels from the training data to fit the interpretable model, since we are now aiming for accuracy, instead of fidelity. Indeed, fitting SFE directly on NELL186 yields an interpretable model with over 90% accuracy (see Table 4), a higher value than XKE-TRUE.

Table 3. Explanations by XKE-TRUE and XKE-PRED (5-NN) in FB13. "cod" and "nat" are abbreviations for "cause_of_death" and "nationality", respectively.

| ID | #1 (XKE-TRUE) | #2 (XKE-TRUE) | #3 (XKE-TRUE) | #4 (XKE-PRED) | #5 (XKE-PRED) |
|---|---|---|---|---|---|
| **Head** | francis_ii_of_the_two_sicilies | ralph_randles_stewart | philip_iv_of_france | hypatia_of_alexandria | guru_dutt |
| **Relation** | RELIGION | RELIGION | CAUSE OF DEATH | PROFESSION | PROFESSION |
| **Tail** | roman_catholic_church | mormon | animal_attack | scientist | war_correspondent |
| **Reason #1** | (2.456) parents, religion | (-0.788) gender, gender$^{-1}$, religion | (-0.286) parents$^{-1}$, gender, gender$^{-1}$, cod | (3.746) cod, profession | (0.233) cod, cod$^{-1}$, profession |
| **Reason #2** | (1.946) spouse$^{-1}$, religion | – | (0.169) spouse, gender, gender$^{-1}$, cod | – | (0.139) nat, nat$^{-1}$, profession |
| **Reason #3** | (1.913) spouse, religion | – | – | – | – |
| **Bias** | (1.017) | (-0.698) | (-0.710) | (-1.418) | (0.239) |
| **XKE** | 0.999346 | 0.184526 | 0.304245 | 0.911209 | 0.648254 |
| **Embedding** | 1 | 0 | 0 | 1 | 0 |

Table 4. Results for learning a logistic regression in features extracted using SFE and labels from the original dataset.

| Dataset | FB13 | NELL186 |
|---|---|---|
| Accuracy | 56.62 | 90.41 |
| Accuracy (filt. examples with # features $> 0$) | 71.88 | 93.88 |
| Accuracy (weighted by # features) | 78.97 | 97.86 |
| F1 | 51.30 | 88.56 |

### 4.4. Qualitative analysis

Now we discuss some examples of predictions made by both XKE variants. Results presented in this section were selected from FB13 in order to represent a variety of cases that the user might face, so as to demonstrate how one can analyze the output of each method. Input triples are shown in Table 3, alongside with their explanations (weighted rules and the bias term), with the interpretable classifier's score, and with the labels predicted by the embedding model.

Examples #1 and #2 are instances of useful explanations that increase or decrease the user's confidence in the model. For both cases, the prediction of XKE-TRUE matches the embedding ones. In #1, the reasons why the religion of `francis_ii_of_the_two_sicilies` is `roman_catholic_church` make sense in practice (i.e., because his parents and his spouse have the same religion as him). These are the sort of correlations that we want our embedding model to capture. For #2 the explanations are also useful, but now in slightly decreasing our confidence in the black box: the path `-gender-gender`$^{-1}$`-religion-` (i.e., a person of the same gender has the religion in question) that explains its low score is not a valid rationale. Ideally, we would expect it to have a coefficient that is closer to zero.

Example #4 is the case where, although XKE-PRED's prediction matches the original classifier's, we can see that the embedding's internal structure makes no sense. The rule learned, with a high coefficient, establishes that the entity that caused `hypatia_of_alexandria`'s death is also a scientist; clearly a type inconsistency.

Finally, examples #3 and #5 are instances where the interpretable classifier is not capable of elucidating the given prediction. In particular, example #5 shows that the interpretable method captured a pertinent correlation that exists in the black box's internal structure (i.e., a person is likely to have had a profession if another person that died for the same reason had the same profession), but that is not useful in explaining this example since the predictions diverge.

## 5. Conclusion

In this paper we proposed two pedagogical methods for explaining embedding models in knowledge bases. Both explain predictions from a black box model in terms of weighted paths in a graph; each path corresponds to a Horn rule that can be interpreted by users. The first method, XKE-PRED, uses other predictions from the embedding model as reference for its explanations. The second method, XKE-TRUE, uses an external source of information, regarded as ground truth, to provide explainable features. Both techniques were able to generate a relatively small number of short weighted Horn clauses that are much easier for a human to interpret than the original embedding space. We expect this initial work to serve as a basis of comparison and inspiration for the development of novel methods for explaining embedding models in KB completion.

The percentage of examples that have no features extracted, and the attained fidelity, are points for improvement. For future work, possible approaches could be to extend the methods presented here by either using additional features that SFE allows for to try to increase the number of examples with at least one feature, or by exploring ways of generating local explanations, aiming for higher fidelity.

## Acknowledgements

The first author is supported by an agreement between IBM and the São Paulo Research Foundation (FAPESP) (grant 2017/19007-6). The fourth author is partially supported by CNPq (grant 308433/2014-9) and received financial support from FAPESP (grant 2016/18841-0).

## References

Andrews, Robert, Diederich, Joachim, and Tickle, Alan B. Survey and critique of techniques for extracting rules from trained artificial neural networks. *Knowledge-Based Systems*, 8(6):373–389, December 1995. ISSN 09507051. doi: 10.1016/0950-7051(96)81920-4.

Augasta, M. Gethsiyal and Kathirvalavakumar, T. Reverse Engineering the Neural Networks for Rule Extraction in Classification Problems. *Neural Processing Letters*, 35(2):131–150, April 2012. ISSN 1370-4621, 1573-773X. doi: 10.1007/s11063-011-9207-8.

Barbieri, Nicola, Bonchi, Francesco, and Manco, Giuseppe. Who to follow and why: Link prediction with explanations. pp. 1266–1275. ACM Press, 2014. ISBN 978-1-4503-2956-9. doi: 10.1145/2623330.2623733.

Bollacker, Kurt, Evans, Colin, Paritosh, Praveen, Sturge, Tim, and Taylor, Jamie. Freebase: A collaboratively created graph database for structuring human knowledge. pp. 1247. ACM Press, 2008. ISBN 978-1-60558-102-6. doi: 10.1145/1376616.1376746.

Bordes, Antoine, Usunier, Nicolas, Garcia-Duran, Alberto, Weston, Jason, and Yakhnenko, Oksana. Translating embeddings for modeling multi-relational data. In *Advances in Neural Information Processing Systems*, pp. 2787–2795, 2013.

Carmona, Ivan Sanchez and Riedel, Sebastian. Extracting interpretable models from matrix factorization models. In *Proceedings of the 2015th International Conference on Cognitive Computation: Integrating Neural and Symbolic Approaches-Volume 1583*, pp. 78–84. CEUR-WS. org, 2015.

Chandrahas, Sengupta, Tathagata, Pragadeesh, Cibi, and Talukdar, Partha Pratim. Inducing Interpretability in Knowledge Graph Embeddings. *arXiv:1712.03547 [cs]*, December 2017.

Craven, Mark and Shavlik, Jude W. Extracting tree-structured representations of trained networks. In *Advances in Neural Information Processing Systems*, pp. 24–30, 1996.

Cucerzan, Silviu. Large-Scale Named Entity Disambiguation Based on Wikipedia Data. In *Proceedings of the 2007 Joint Conference on Empirical Methods in Natural Language Processing and Computational Natural Language Learning (EMNLP-CoNLL)*, pp. 708–716, Prague, Czech Republic, June 2007. Association for Computational Linguistics.

d'Avila Garcez, A., Besold, Tarek R., De Raedt, Luc, Földiak, Peter, Hitzler, Pascal, Icard, Thomas, Kühnberger, Kai-Uwe, Lamb, Luis C., Miikkulainen, Risto, and Silver, Daniel L. Neural-symbolic learning and reasoning: Contributions and challenges. In *Proceedings of the AAAI Spring Symposium on Knowledge Representation and Reasoning: Integrating Symbolic and Neural Approaches, Stanford*, 2015.

Doran, Derek, Schulz, Sarah, and Besold, Tarek R. What Does Explainable AI Really Mean? A New Conceptualization of Perspectives. *arXiv:1710.00794 [cs]*, October 2017.

Engelen, Jesper, Boekhout, Hanjo, and Takes, Frank. Explainable and Efficient Link Prediction in Real-World Network Data. volume 9897 of *Lecture Notes in Computer Science*, Cham, 2016. Springer International Publishing. ISBN 978-3-319-46348-3 978-3-319-46349-0.

Etchells, T.A. and Lisboa, P.J.G. Orthogonal Search-Based Rule Extraction (OSRE) for Trained Neural Networks: A Practical and Efficient Approach. *IEEE Transactions on Neural Networks*, 17(2):374–384, March 2006. ISSN 1045-9227. doi: 10.1109/TNN.2005.863472.

França, Manoel Vitor Macedo, D'Avila Garcez, Artur S., and Zaverucha, Gerson. Relational knowledge extraction from neural networks. In *Proceedings of the 2015th International Conference on Cognitive Computation: Integrating Neural and Symbolic Approaches-Volume 1583*, pp. 146–154. CEUR-WS. org, 2015.

Freitas, Alex A. Comprehensible Classification Models – a position paper. *ACM SIGKDD Explorations*, 15(1):10, 2013.

Gardner, Matt and Mitchell, Tom M. Efficient and Expressive Knowledge Base Completion Using Subgraph Feature Extraction. In *EMNLP*, pp. 1488–1498, 2015.

Gardner, Matt, Talukdar, Partha, and Mitchell, Tom. Combining Vector Space Embeddings with Symbolic Logical Inference over Open-Domain Text. pp. 5, 2015.

Guo, Shu, Wang, Quan, Wang, Bin, Wang, Lihong, and Guo, Li. Semantically Smooth Knowledge Graph Embedding. In *Proceedings of the 53rd Annual Meeting of the Association for Computational Linguistics and the 7th International Joint Conference on Natural Language*

*Processing (Volume 1: Long Papers)*, pp. 84–94, Beijing, China, July 2015. Association for Computational Linguistics.

Lao, Ni and Cohen, William W. Relational retrieval using a combination of path-constrained random walks. *Machine Learning*, 81(1):53–67, October 2010. ISSN 0885-6125, 1573-0565. doi: 10.1007/s10994-010-5205-8.

Lao, Ni, Mitchell, Tom, and Cohen, William W. Random walk inference and learning in a large scale knowledge base. In *Proceedings of the Conference on Empirical Methods in Natural Language Processing*, pp. 529–539. Association for Computational Linguistics, 2011.

Lipton, Zachary C. The Mythos of Model Interpretability. In *Proceedings of the ICML Workshop on Human Interpretability in Machine Learning*, pp. 96–100, June 2016.

Liu, Hanxiao, Wu, Yuexin, and Yang, Yiming. Analogical Inference for Multi-Relational Embeddings. *arXiv:1705.02426 [cs]*, May 2017.

Mikolov, Tomas, Chen, Kai, Corrado, Greg, and Dean, Jeffrey. Efficient Estimation of Word Representations in Vector Space. *arXiv:1301.3781 [cs]*, January 2013.

Miller, George A. WordNet: A lexical database for English. *Commun. ACM*, pp. 39–41, November 1995. doi: 10.1145/219717.219748.

Mitchell, Tom M., Cohen, William W., Hruschka Jr, Estevam R., Talukdar, Partha Pratim, Betteridge, Justin, Carlson, Andrew, Mishra, Bhavana Dalvi, Gardner, Matthew, Kisiel, Bryan, Krishnamurthy, Jayant, and others. Never Ending Learning. In *AAAI*, pp. 2302–2310, 2015.

Murphy, Brian, Talukdar, Partha, and Mitchell, Tom. Learning effective and interpretable semantic models using non-negative sparse embedding. *Proceedings of COLING 2012*, pp. 1933–1950, 2012.

Nguyen, Dat Quoc. An overview of embedding models of entities and relationships for knowledge base completion. *arXiv:1703.08098 [cs]*, March 2017.

Nguyen, Dat Quoc, Sirts, Kairit, Qu, Lizhen, and Johnson, Mark. Neighborhood Mixture Model for Knowledge Base Completion. *arXiv:1606.06461 [cs]*, pp. 40–50, 2016. doi: 10.18653/v1/K16-1005.

Nickel, Maximilian, Murphy, Kevin, Tresp, Volker, and Gabrilovich, Evgeniy. A Review of Relational Machine Learning for Knowledge Graphs. *Proceedings of the IEEE*, 104(1):11–33, 2015. ISSN 0018-9219, 1558-2256. doi: 10.1109/JPROC.2015.2483592.

Ribeiro, Marco Tulio, Singh, Sameer, and Guestrin, Carlos. "Why Should I Trust You?": Explaining the Predictions of Any Classifier. *arXiv:1602.04938 [cs, stat]*, February 2016.

Schuhmacher, Michael and Ponzetto, Simone Paolo. Knowledge-based graph document modeling. pp. 543–552. ACM Press, 2014. ISBN 978-1-4503-2351-2. doi: 10.1145/2556195.2556250.

Socher, Richard, Chen, Danqi, Manning, Christopher D., and Ng, Andrew. Reasoning with neural tensor networks for knowledge base completion. In *Advances in Neural Information Processing Systems*, pp. 926–934, 2013.

Srinivasan, Ashwin and Vig, Lovekesh. Mode-Directed Neural-Symbolic Modelling. In *Submitted to: The 27th International Conference on Inductive Logic Programming (ILP2017)*, 2017.

Tintarev, Nava and Masthoff, Judith. A survey of explanations in recommender systems. In *Data Engineering Workshop, 2007 IEEE 23rd International Conference On*, pp. 801–810. IEEE, 2007.

Trouillon, Théo, Welbl, Johannes, Riedel, Sebastian, Gaussier, Éric, and Bouchard, Guillaume. Complex Embeddings for Simple Link Prediction. *arXiv:1606.06357 [cs, stat]*, June 2016.

Wang, Q., Mao, Z., Wang, B., and Guo, L. Knowledge Graph Embedding: A Survey of Approaches and Applications. *IEEE Transactions on Knowledge and Data Engineering*, 29(12):2724–2743, December 2017. ISSN 1041-4347. doi: 10.1109/TKDE.2017.2754499.

Wang, Zhen, Zhang, Jianwen, Feng, Jianlin, and Chen, Zheng. Knowledge Graph Embedding by Translating on Hyperplanes. pp. 8, 2014.

Xiao, Han. KSR: A Semantic Representation of Knowledge Graph within a Novel Unsupervised Paradigm. *arXiv:1608.07685 [cs]*, August 2016.

Xie, Qizhe, Ma, Xuezhe, Dai, Zihang, and Hovy, Eduard. An Interpretable Knowledge Transfer Model for Knowledge Base Completion. *arXiv:1704.05908 [cs]*, 2017.

Zilke, Jan Ruben, Mencía, Eneldo Loza, and Janssen, Frederik. DeepRED – Rule Extraction from Deep Neural Networks. volume 9956 of *Lecture Notes in Computer Science*, Cham, 2016. Springer International Publishing. ISBN 978-3-319-46306-3 978-3-319-46307-0. doi: 10.1007/978-3-319-46307-0_29.